\documentclass[11pt]{article}
\usepackage{acl2016}
\usepackage{times}
\usepackage{latexsym}
\usepackage{url}
\usepackage{amsmath}
\usepackage{multirow}
\usepackage{graphicx}
\usepackage{color}
\usepackage{tikz-qtree}
\usepackage{tikz}
\usepackage{tikz-dependency}
\usepackage{sidecap}
\aclfinalcopy

\DeclareMathOperator*{\argmax}{arg\,max}
\newcommand{\rpm}{\raisebox{.2ex}{$\scriptstyle\pm$}}
\newcommand{\ignore}[1]{}

\newcommand{\sect}[1]{\S\ref{#1}}
\newcommand{\figu}[1]{Figure~\ref{#1}}
\newcommand{\tabr}[1]{Table~\ref{#1}}

\newcommand{\rt}{\textsf{root}}

\newcommand{\specialcell}[2][l]{%
  \begin{tabular}[#1]{@{}l@{}r}#2\end{tabular}}

\usepackage{enumitem}

\setitemize{noitemsep,topsep=6pt,parsep=0pt,partopsep=0pt,itemindent=0pt,leftmargin=12pt,labelwidth=12pt}
\setenumerate{noitemsep,topsep=6pt,parsep=0pt,partopsep=0pt,itemindent=0pt,leftmargin=12pt,labelwidth=12pt}

\title{Greedy, Joint Syntactic-Semantic Parsing with Stack LSTMs}

\author{Swabha Swayamdipta$^{\clubsuit}$ ~ Miguel Ballesteros$^{\diamondsuit}$ ~ Chris Dyer$^{\spadesuit}$ ~ Noah A. Smith$^{\heartsuit}$\\
  $^\clubsuit$School of Computer Science, Carnegie Mellon University,
  Pittsburgh, PA 15213, USA\\
$^\diamondsuit$Natural Language Processing Group,
  Universitat Pompeu Fabra, Barcelona, Spain\\  $^{\spadesuit}$Google
  DeepMind, London, UK \\
$^\heartsuit$Computer Science \& Engineering, University of
Washington, Seattle, WA 98195, USA
\\
 {\tt swabha@cs.cmu.edu, miguel.ballesteros@upf.edu,} \\
  {\tt cdyer@cs.cmu.edu, nasmith@cs.washington.edu} \\ 
  }
\date{}

\begin{document}
\maketitle
\begin{abstract}
We present a transition-based parser that jointly produces syntactic and semantic dependencies. It learns a representation of the entire
algorithm state, using stack long short-term memories. Our greedy
inference algorithm has linear time, including feature extraction. On
the CoNLL 2008--9 English shared tasks, we obtain the best published parsing performance among models that jointly learn syntax and semantics.

\end{abstract}

\section{Introduction} 

We introduce a new joint syntactic and semantic dependency parser. Our parser draws from the algorithmic insights of the incremental structure building approach of \newcite{Henderson:08}, with two key differences. First, it learns representations for the parser's entire algorithmic state, not just the top items on the stack or the most recent parser states; in fact, it uses no expert-crafted features at all. Second, it uses entirely greedy inference rather than beam search. We find that it outperforms all previous joint parsing models, including~\newcite{Henderson:08} and variants \cite{Gesmundo:09,Titov:09,Henderson:13} on the CoNLL 2008 and 2009 (English) shared tasks. Our parser's multilingual results are comparable to the top systems at CoNLL 2009.

Joint models like ours have frequently been proposed as a way to avoid cascading errors in NLP pipelines; varying degrees of success have been attained for a range of joint syntactic-semantic analysis tasks \cite[\em inter alia]{Sutton:05,Henderson:08,Toutanova:08,Johansson:09,Lluis:13}.

One reason pipelines often dominate is that they make available the complete syntactic parse tree, and arbitrarily-scoped syntactic features---such as the ``path'' between predicate and argument, proposed by \newcite{Gildea:02a}---for semantic analysis. Such features are a mainstay of high-performance semantic role labeling (SRL) systems \cite{Roth:14,Lei:15,Fitzgerald:15,Foland:15}, but they are expensive to extract \cite{Johansson:09,He:13}. 

This study shows how recent advances in representation learning can bypass those expensive features, discovering cheap alternatives available during a greedy parsing procedure. The specific advance we employ is the stack LSTM \cite{Dyer:15}, a neural network that continuously summarizes the contents of the stack data structures in which a transition-based parser's state is conventionally encoded. Stack LSTMs were shown to obviate many features used in syntactic dependency parsing; here we find them to do the same for joint syntactic-semantic dependency parsing. 

We believe this is an especially important finding for \emph{greedy} models that cast parsing as a sequence of decisions made based on algorithmic state, where linguistic theory and researcher intuitions offer less guidance in feature design.

Our system's performance does not match that of the top expert-crafted feature-based systems \cite{Zhao:09,Bjorkelund:10,Roth:14,Lei:15}, systems which perform optimal decoding~\cite{Tackstrom:15}, or of systems that exploit additional, differently-annotated datasets~\cite{Fitzgerald:15}. Many advances in those systems are orthogonal to our model, and we expect future work to achieve further gains by integrating them. 

Because our system is very fast---
with an end-to-end runtime of $177.6 \rpm 18$ seconds to parse the CoNLL 2009 English test data on a single core---we believe it will be useful in practical settings. Our open-source implementation has been released.\footnote{\url{https://github.com/clab/joint-lstm-parser}}

\begin{figure}
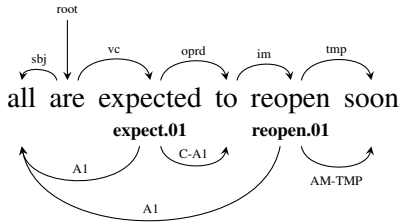

	\begin{center}
		\begin{dependency} [theme = simple]
			\label{dep1}
			\begin{deptext}
				all \& are \& expected \& to \& reopen \& soon \& \\
				\& \& \scriptsize{\textbf{expect.01}} \& \& \scriptsize{\textbf{reopen.01}} \& \& \\
			\end{deptext}
			\depedge{2}{1}{sbj}
			\deproot[edge unit distance=1.5ex]{2}{root}
			\depedge{2}{3}{vc}
			\depedge{3}{4}{oprd}
			\depedge{4}{5}{im}
			\depedge{5}{6}{tmp}
			\depedge[edge below]{3}{1}{A1}
			\depedge[edge below]{3}{4}{C-A1}
			\depedge[edge below, label style={below}]{5}{6}{AM-TMP}
			\depedge[edge below]{5}{1}{A1}
		\end{dependency}
	\end{center}
	\begin{scriptsize}
		\caption{Example of a joint parse. Syntactic dependencies are shown by arcs above the sentence and semantic dependencies below; predicates are marked in boldface. C- denotes continuation of argument A1. Correspondences between dependencies might be close (between \emph{expected} and \emph{to}) or not (between \emph{reopen} and \emph{all}). 
			\label{fig:illust}}
	\end{scriptsize}
\end{figure}

\section{Joint Syntactic and Semantic Dependency Parsing}
\label{sec:algo}

We largely follow the transition-based, synchronized algorithm of \newcite{Henderson:13} to predict joint parse structures. The input to the algorithm is a sentence annotated with part-of-speech tags. The output consists of a labeled syntactic dependency tree and a directed SRL graph, in which a subset of words in the sentence are selected as predicates, disambiguated to a sense, and linked by labeled, directed edges to their semantic arguments. \figu{fig:illust} shows an example.

\subsection{Transition-Based Procedure}

The two parses are constructed in a bottom-up fashion, incrementally processing words in the sentence from left to right. The state of the parsing algorithm at timestep $t$ is represented by three stack data structures: a syntactic stack $S_t$, a semantic stack $M_t$---each containing partially built structures---and a buffer of input words $B_t$. Our algorithm also places partial syntactic and semantic parse structures onto the front of the buffer, so it is also implemented as a stack. Each arc in the output corresponds to a transition (or ``action'') chosen based on the current state; every transition modifies the state by updating $S_t$, $M_t$, and $B_t$ to $S_{t+1}$, $M_{t+1}$, and $B_{t+1}$, respectively. While each state may license several valid actions, each action has a deterministic effect on the state of the algorithm.

Initially, $S_0$ and $M_0$ are empty, and $B_0$ contains the input sentence with the first word at the front of $B$ and a special \rt~symbol at the end.\footnote{This works better for the arc-eager algorithm~\cite{Ballesteros:13}, in contrast to \newcite{Henderson:13}, who initialized with \rt~at the buffer front.} Execution ends on iteration $t$ such that $B_t$ is empty and $S_t$ and $M_t$ contain only a single structure headed by \rt. 

\subsection{Transitions for Joint Parsing}
\label{sec:jtrans}

There are separate sets of syntactic and semantic transitions; the former manipulate $S$ and $B$, the latter $M$ and $B$. All are formally defined in \tabr{tab:trans}. The syntactic transitions are from the ``arc-eager'' algorithm of \newcite{Nivre:08}. They include: 
\begin{itemize}
	\item \textsc{S-Shift}, which copies\footnote{Note that in the
		original arc-eager algorithm~\cite{Nivre:08}, \textsc{Shift} and
		\textsc{Right-Arc} actions \emph{move} the item on the buffer front to
		the stack, whereas we only copy it (to allow the semantic
		operations to have access to it).} an item from the front of $B$ and pushes it on $S$.
	\item \textsc{S-Reduce} pops an item from $S$.
	\item \textsc{S-Right}$(\ell)$ creates a syntactic dependency. Let $u$ be the element at the top of $S$ and $v$ be the element at the front of $B$. The new dependency has $u$ as head, $v$ as dependent, and label $\ell$. $u$ is popped off $S$, and the resulting structure, rooted at $u$, is pushed on $S$. Finally, $v$ is copied to the top of $S$.
	\item \textsc{S-Left}$(\ell)$ creates a syntactic dependency with label $\ell$ in the reverse direction as \textsc{S-Right}. The top of $S$, $u$, is popped. The front of $B$, $v$, is replaced by the new structure, rooted at $v$.
\end{itemize}

The semantic transitions are similar, operating on the semantic stack.
\begin{itemize}
	\item \textsc{M-Shift} removes an item from the front of $B$ and pushes it on $M$.
	\item \textsc{M-Reduce} pops an item from $M$.
	\item \textsc{M-Right}$(r)$ creates a semantic dependency. Let $u$
	be the element at the top of $M$ and $v$, the front of $B$. The new dependency has $u$ as
	head, $v$ as dependent, and label $r$. $u$ is popped off $M$, and the resulting structure, rooted at $u$, is
	pushed on $M$. 
	\item \textsc{M-Left}$(r)$ creates a semantic dependency with label
	$r$ in the reverse direction as \textsc{M-Right}. The buffer front, $v$, is replaced by the
	new $v$-rooted structure. $M$ remains unchanged.
\end{itemize}

Because SRL graphs allow a node to be a semantic argument of two parents---like \emph{all} in the example in Figure~\ref{fig:illust}---\textsc{M-Left} and \textsc{M-Right} do not remove the dependent from the semantic stack and buffer respectively, unlike their syntactic equivalents, \textsc{S-Left} and \textsc{S-Right}. We use two other semantic transitions from~\newcite{Henderson:13} which have no syntactic analogues:
\begin{itemize}
	\item \textsc{M-Swap} swaps the top two items on $M$, to allow for crossing semantic arcs.
	\item \textsc{M-Pred}$(p)$ marks the item at the front of $B$ as a semantic predicate with the sense $p$, and replaces it with the disambiguated predicate. 
\end{itemize}

\begin{figure}
	\begin{center}
		\includegraphics[width=0.5\textwidth]{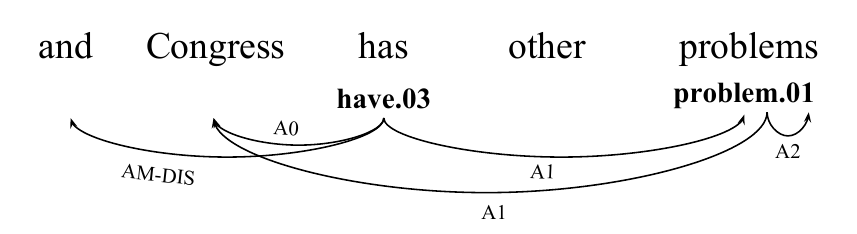}
	\end{center}
	\caption{Example of an SRL graph with an arc from predicate \textbf{problem.01} to itself, filling the A2 role. Our \textsc{Self}(A2) transition allows recovering this semantic dependency. 
		\label{fig:self}}
\end{figure}

The CoNLL 2009 corpus introduces semantic self-dependencies where many nominal predicates (from NomBank) are marked as their own arguments; these account for 6.68\% of all semantic arcs in the English corpus. An example involving an eventive noun is shown in~\figu{fig:self}. We introduce a new semantic transition, not in~\newcite{Henderson:13}, to handle such cases:
\begin{itemize}
	\item \textsc{M-Self}$(r)$ adds a dependency, with label $r$ between the item at the front of $B$ and itself. The result replaces the item at the front of $B$.
\end{itemize}

\begin{table*}
	\centering\small
	\begin{scriptsize}
		\begin{tabular}{ccc|l|crc|c}
			$S_t$ & $M_t$ & $B_t$ & \textbf{Action} & $S_{t+1}$ & $M_{t+1}$ & $B_{t+1}$ & \textbf{Dependency} \\
			\hline
			$S$ & $M$ & $(\mathbf{v},v),B$ & \textsc{S-Shift} & $(\mathbf{v},v),S$ & $M$ & $(\mathbf{v},v),B$ & --- \\
			$(\mathbf{u},u),S$ & $M$ & $B$ & \textsc{S-Reduce} & $S$ & $M$ & $B$ & --- \\
			$(\mathbf{u},u),S$ & $M$ & $(\mathbf{v},v),B$ & \textsc{S-Right}$(\ell)$ &\multicolumn{2}{l}{$(\mathbf{v},v), (g_s(\mathbf{u},\mathbf{v},\mathbf{l}),u), S$ \hfill $M$} & $(\mathbf{v},v),B$ & $\mathcal{S} \cup u \stackrel{\scriptsize{\ell}}{\rightarrow} v$\\
			$(\mathbf{u},u),S$ & $M$ & $(\mathbf{v},v),B$ & \textsc{S-Left}$(\ell)$ & $S$ & $M$ & $(g_s(\mathbf{v},\mathbf{u},\mathbf{l}), v),B$ & $\mathcal{S} \cup u \stackrel{\scriptsize{\ell}}{\leftarrow} v$ \\ \hline
			
			$S$ & $M$ & $(\mathbf{v},v),B$ & \textsc{M-Shift} & $S$ & $(\mathbf{v},v),M$ & $B$ & --- \\
			$S$ & $(\mathbf{u},u),M$ & $B$ & \textsc{M-Reduce} & $S$ & $M$ & $B$ & --- \\
			$S$ & $(\mathbf{u},u),M$ & $(\mathbf{v},v),B$ & \textsc{M-Right}$(r)$ & $S$ & $(g_m(\mathbf{u},\mathbf{v}, \mathbf{r}),u),M$ & $(\mathbf{v},v),B$ & $\mathcal{M} \cup u \stackrel{\scriptsize{r}}{\rightarrow} v$ \\
			$S$ & $(\mathbf{u},u),M$ & $(\mathbf{v},v),B$ & \textsc{M-Left}$(r)$ & $S$ & $(\mathbf{u},u),M$ & $(g_m(\mathbf{v},\mathbf{u}, \mathbf{r}),v), B$ & $\mathcal{M} \cup u \stackrel{\scriptsize{r}}{\leftarrow} v$ \\
			$S$ & $(\mathbf{u},u),(\mathbf{v},v),M$ & $B$ & \textsc{M-Swap} & $S$ & $(\mathbf{v},v),(\mathbf{u},u), M$ & $B$ & --- \\
			$S$ & $M$ & $(\mathbf{v},v),B$ & \textsc{M-Pred}$(p)$ & $S$ & $M$ & $(g_d(\mathbf{v}, \mathbf{p}), v) ,B$ & --- \\
			$S$ & $M$ & $(\mathbf{v},v),B$ & \textsc{M-Self}$(r)$ & $S$ & $M$ & $(g_m(\mathbf{v},\mathbf{v}, \mathbf{r}),v), B$ & $\mathcal{M} \cup v \stackrel{\scriptsize{r}}{\leftrightarrow} v$ \\
		\end{tabular}
	\end{scriptsize}
	\caption{\label{tab:trans}Parser transitions along with the
		modifications to the stacks and the buffer resulting from each. Syntactic
		transitions are shown above, semantic below. Italic symbols denote symbolic representations of words and
		relations, and bold symbols indicate (learned) embeddings (\sect{sec:pretrained}) of words and
		relations; each element in a stack or buffer includes both symbolic
		and vector representations, either atomic or recursive. $\mathcal{S}$ represents the set of syntactic transitions, and $\mathcal{M}$ the set of semantic transitions. }
\end{table*}

Note that the syntactic and semantic transitions both operate on the same buffer, though they independently specify the syntax and semantics, respectively. In order to ensure that both syntactic and semantic parses are produced, the syntactic and semantic transitions are interleaved. Only syntactic transitions are considered until a transition is chosen that copies an item from the buffer front to the syntactic stack (either \textsc{S-Shift} or \textsc{S-Right}). The algorithm then switches to semantic transitions until a buffer-modifying transition is taken (\textsc{M-Shift}).\footnote{Had we \textit{moved} the item at the buffer front during the syntactic transitions, it would have been unavailable for the semantic transitions, hence we only \textit{copy} it.} At this point, the buffer is modified and the algorithm returns to syntactic transitions. This implies that, for each word, its left-side syntactic dependencies are resolved before its left-side semantic dependencies. An example run of the algorithm is shown in \figu{fig:ex}.

\subsection{Constraints on Transitions}
\label{sec:constraints}

To ensure that the parser never enters an invalid state, the sequence of transitions is constrained, following \newcite{Henderson:13}. Actions that copy or move items from the buffer (\textsc{S-Shift}, \textsc{S-Right} and \textsc{M-Shift}) are forbidden when the buffer is empty. Actions that pop from a stack (\textsc{S-Reduce} and \textsc{M-Reduce}) are forbidden when that stack is empty. We disallow actions corresponding to the same dependency, or the same predicate to be repeated in the sequence. Repetitive \textsc{M-Swap} transitions are disallowed to avoid infinite swapping. Finally, as noted above, we restrict the parser to syntactic actions until it needs to shift an item from $B$ to $S$, after which it can only execute semantic actions until it executes an \textsc{M-Shift}.

Asymptotic runtime complexity of this greedy algorithm is linear in the length of the input, following the analysis by \newcite{Nivre:09}.\footnote{The analysis in \cite{Nivre:09} does not consider \textsc{Swap} actions. However, since we constrain the number of such actions, the linear time complexity of the algorithm stays intact.}

\begin{figure*}[!ht]
	\begin{center}
		\centering
		\scalebox{0.8}{
			\rotatebox{90}{
				\begin{tabular}{lllll}\textbf{Transition}&$S$&$M$&$B$&\textbf{Dependency}\\
					\hline
					&[]&[]&[all, are, expected, to, reopen, soon, \rt]& --- \\
					\textsc{S-Shift}& [all]& []& [all, are, expected, to, reopen, soon, \rt]& --- \\
					\textsc{M-Shift}& [all]& [all]& [are, expected, to, reopen, soon, \rt]& --- \\
					\textsc{S-Left}(\textit{sbj})& []& [all]& [are, expected, to, reopen, soon, \rt]& all $\stackrel{\scriptsize{\textrm{sbj}}}{\longleftarrow}$ are \\
					\textsc{S-Shift}& [are]& [all]& [are, expected, to, reopen, soon, \rt]& --- \\
					\textsc{M-Shift}& [are]& [all, are]& [expected, to, reopen, soon, \rt]& --- \\
					\textsc{S-Right}(\textit{vc})& [are, expected]& [all, are]& [expected, to, reopen, soon, \rt]& are $\stackrel{\scriptsize{\textrm{vc}}}{\longrightarrow}$ expected \\
					\textsc{M-Pred}(\textbf{expect.01}) & [are, expected]& [all, are]& [expected, to, reopen, soon, \rt]& --- \\
					\textsc{M-Reduce}& [are, expected]& [all]& [expected, to, reopen, soon, \rt]& --- \\
					\textsc{M-Left}(\textit{A1}) & [are, expected]& [all]& [expected, to, reopen, soon, \rt]& all $\stackrel{\scriptsize{\textrm{A1}}}{\longleftarrow}$ expect.01 \\
					\textsc{M-Shift}& [are, expected]& [all, expected]& [to, reopen, soon, \rt]& --- \\
					***\textsc{S-Right}(oprd) & [are, expected, to]& [all, expected]& [to, reopen, soon, \rt]& expected $ \stackrel{\scriptsize{\textrm{oprd}}}{\longrightarrow} $ to \\
					\textsc{M-Right}(\textit{C-A1}) & [are, expected, to]& [all, expected]& [to, reopen, soon, \rt]& expect.01 $ \stackrel{\scriptsize{\textrm{C-A1}}}{\longrightarrow} $ to \\
					\textsc{M-Reduce}& [are, expected, to]& [all]& [to, reopen, soon, \rt]& --- \\
					\textsc{M-Shift}& [are, expected, to]& [all, to]& [reopen, soon, \rt]& --- \\
					\textsc{S-Right}(\textit{im}) & [are, expected, to, reopen]& [all, to]& [reopen, soon, \rt]& to $ \stackrel{\scriptsize{\textrm{im}}}{\longrightarrow} $ reopen\\
					\textsc{M-Pred}(\textbf{reopen.01}) & [are, expected, to, reopen]& [all, to]& [reopen, soon, \rt]& --- \\
					\textsc{M-Reduce}& [are, expected, to, reopen]& [all]& [reopen, soon, \rt]& --- \\
					\textsc{M-Left}(\textit{A1}) & [are, expected, to, reopen]& [all]& [reopen, soon, \rt]& all $ \stackrel{\scriptsize{\textrm{A1}}}{\longleftarrow} $ reopen.01 \\
					\textsc{M-Reduce}& [are, expected, to, reopen]& []& [reopen, soon, \rt]& --- \\
					\textsc{M-Shift}& [are, expected, to, reopen]& [reopen]& [soon, \rt]& --- \\
					\textsc{S-Right}(\textit{tmp})& [are, expected, to, reopen, soon]& [reopen]& [soon, \rt]& reopen $ \stackrel{\scriptsize{\textrm{tmp}}}{\longrightarrow} $ soon \\
					\textsc{M-Right}(\textit{AM-TMP}) & [are, expected, to, reopen, soon]& [reopen]& [soon, \rt]& reopen.01 $ \stackrel{\textrm{AM-TMP}}{\longrightarrow} $ soon \\
					\textsc{M-Reduce}& [are, expected, to, reopen, soon]& []& [soon, \rt]& --- \\
					\textsc{M-Shift}& [are, expected, to, reopen, soon]& [soon]& [\rt]& --- \\
					\textsc{S-Reduce}& [are, expected, to, reopen]& [soon]& [\rt]& --- \\
					\textsc{S-Reduce}& [are, expected, to]& [soon]& [\rt]& --- \\
					\textsc{S-Reduce}& [are, expected]& [soon]& [\rt]& --- \\
					\textsc{S-Reduce}& [are]& [soon]& [\rt]& --- \\
					\textsc{S-Left}(\textit{root}) & []& [soon]& [\rt]& are $\stackrel{\scriptsize{\textrm{root}}}{\longleftarrow}$ \rt \\
					\textsc{S-Shift}& [\rt]& [soon]& [\rt]& --- \\
					\textsc{M-Reduce}& [\rt]& []& [\rt]& --- \\
					\textsc{M-Shift}& [\rt]& [\rt]& []& --- \\
			\end{tabular}}
		}
		\caption{Joint parser transition sequence for the sentence in \ignore{Fig. 1}\figu{fig:illust}, ``\emph{all are expected to reopen soon}.'' Syntactic labels are in lower-case and semantic role labels are capitalized.
			*** marks the operation predicted in \figu{fig:model}\ignore{Fig. 3}.
			\label{fig:ex} }
	\end{center}
\end{figure*}

\section{Statistical Model}
\label{sec:model}

The transitions in \S\ref{sec:algo} describe the execution paths our algorithm can take; like past work, we apply a statistical classifier to decide which transition to take at each timestep, given the current state. The novelty of our model is that it learns a finite-length vector representation of the entire joint parser's state ($S$, $M$, and $B$) in order to make this decision.

\subsection{Stack Long Short-Term Memory (LSTM)}
\label{sec:slstm}

LSTMs are recurrent neural networks equipped with specialized memory components in addition to a hidden state~\cite{Hochreiter:97,Graves:13} to model sequences. Stack LSTMs~\cite{Dyer:15} are LSTMs that allow for stack operations: \emph{query}, \emph{push}, and \emph{pop}. A ``stack pointer'' is maintained which determines which cell in the LSTM provides the memory and hidden units when computing the new memory cell contents. \emph{Query} provides a summary of the stack in a single fixed-length vector. \emph{Push} adds an element to the top of the stack, resulting in a new summary.  \emph{Pop}, which does not correspond to a conventional LSTM operation, moves the stack pointer to the preceding  timestep, resulting in a stack summary as it was before the popped item was observed. Implementation details~\cite{Dyer:15,Goldberg:15} and code have been made publicly available.\footnote{\url{https://github.com/clab/lstm-parser}}

Using stack LSTMs, we construct a representation of the algorithm state by decomposing it into smaller pieces that are combined by recursive function evaluations (similar to the way a list is built by a \emph{concatenate} operation that operates on a list and an element). This enables information that would be distant from the ``top'' of the stack to be carried forward, potentially helping the learner.


\subsection{Stack LSTMs for Joint Parsing}
\label{sec:jointslstm}

\begin{figure}
	\begin{center}
		\includegraphics[width=0.5\textwidth]{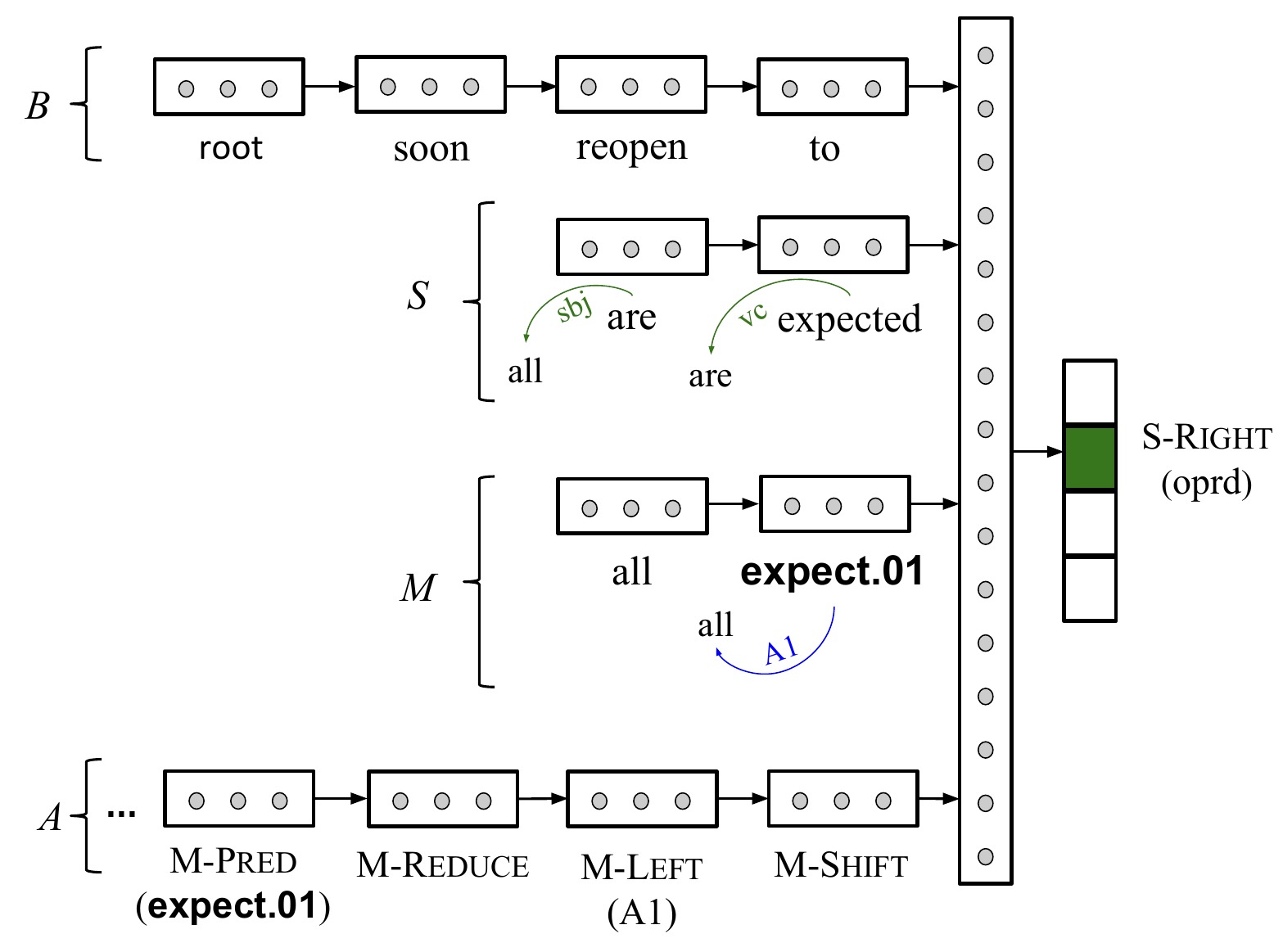}
	\end{center}
	\caption{Stack LSTM for joint parsing. The state illustrated corresponds to the ***-marked row in the example transition sequence in Fig.~\ref{fig:ex}.
		\label{fig:model}}
\end{figure}

Our algorithm employs four stack LSTMs, one each for the $S$, $M$, and $B$ data structures.
Like~\newcite{Dyer:15}, we use a fourth stack LSTM, $A$, for the history of actions---$A$ is never popped from, only pushed to. \figu{fig:model} illustrates the architecture. 
The algorithm's state at timestep $t$ is encoded by the four vectors summarizing the four stack LSTMs, and this is the input to the classifier that chooses among the allowable transitions at that timestep. 

Let $\mathbf{s}_t$, $\mathbf{m}_t$, $\mathbf{b}_t$, and $\mathbf{a}_t$
denote the summaries of $S_t$, $M_t$, $B_t$, and $A_t$, respectively. Let $\mathcal{A}_t = \mathrm{Allowed}(S_t, M_t, B_t, A_t)$ denote the allowed transitions given the current stacks and buffer. The parser state at time $t$ is given by a rectified linear unit~\cite{Nair:10} in vector $\mathbf{y}_t$:
\begin{align}
\mathbf{y}_t =& \ \mathrm{elementwisemax}\left\{ \boldsymbol{0}, \mathbf{d} + \mathbf{W} [\mathbf{s}_t ; \mathbf{m}_t ; \mathbf{b}_t; \mathbf{a}_t] \right\} \nonumber
\end{align}
where $\mathbf{W}$ and $\mathbf{d}$ are the parameters of the classifier. The transition selected at timestep $t$ is 
\begin{align}
& \argmax_{\tau \in \mathcal{A}_t} q_\tau + \boldsymbol{\theta}_\tau \cdot \mathbf{y}_t \label{eq:decision} \\
\equiv& \argmax_{\tau \in \mathcal{A}_t} \mathrm{score}(\tau; S_t, M_t, B_t, A_t) \nonumber
\end{align}

\noindent where $\boldsymbol{\theta}_\tau$ and $q_\tau$ are parameters for each transition type $\tau$. Note that only allowed transitions are considered in the decision rule (see~\S\ref{sec:constraints}). 

\subsection{Composition Functions}
To use stack LSTMs, we require vector representations of the elements 
that are stored in the stacks. Specifically, we require vector
representations of atoms (words, possibly with part-of-speech tags)
and parse fragments. Word vectors can be pretrained or learned
directly; we consider a concatenation of both in our experiments; part-of-speech vectors are learned and concatenated to the same. 

To obtain vector representations of parse fragments, we use neural networks which recursively compute representations of the complex structured output~\cite{Dyer:15}. The tree structures here are always ternary trees, with each internal node's three children including a head, a dependent, and a label. The vectors for leaves are word vectors and vectors corresponding to syntactic and semantic relation types.

The vector for an internal node is a squashed ($\tanh$) affine transformation of its children's vectors. For syntactic and semantic attachments, respectively, the composition function is:
\begin{align}
g_{s}(\mathbf{v}, \mathbf{u}, \mathbf{l}) =& \tanh(\mathbf{Z}_s [\mathbf{v}; \mathbf{u}; \mathbf{l}] + \mathbf{e}_s)\\
g_{m}(\mathbf{v}, \mathbf{u}, \mathbf{r}) =& \tanh(\mathbf{Z}_m [\mathbf{v}; \mathbf{u}; \mathbf{r}] + \mathbf{e}_m)
\end{align}
where $\mathbf{v}$ and $\mathbf{u}$ are vectors corresponding to
atomic words or composed parse fragments; $\mathbf{l}$ and $\mathbf{r}$ are learned vector
representations for syntactic and semantic labels respectively. Syntactic and semantic
parameters are separated ($\mathbf{Z}_s$, $\mathbf{e}_s$ and $\mathbf{Z}_m$, $\mathbf{e}_m$, respectively).

\begin{figure}
	\centering
	\includegraphics[width=\linewidth]{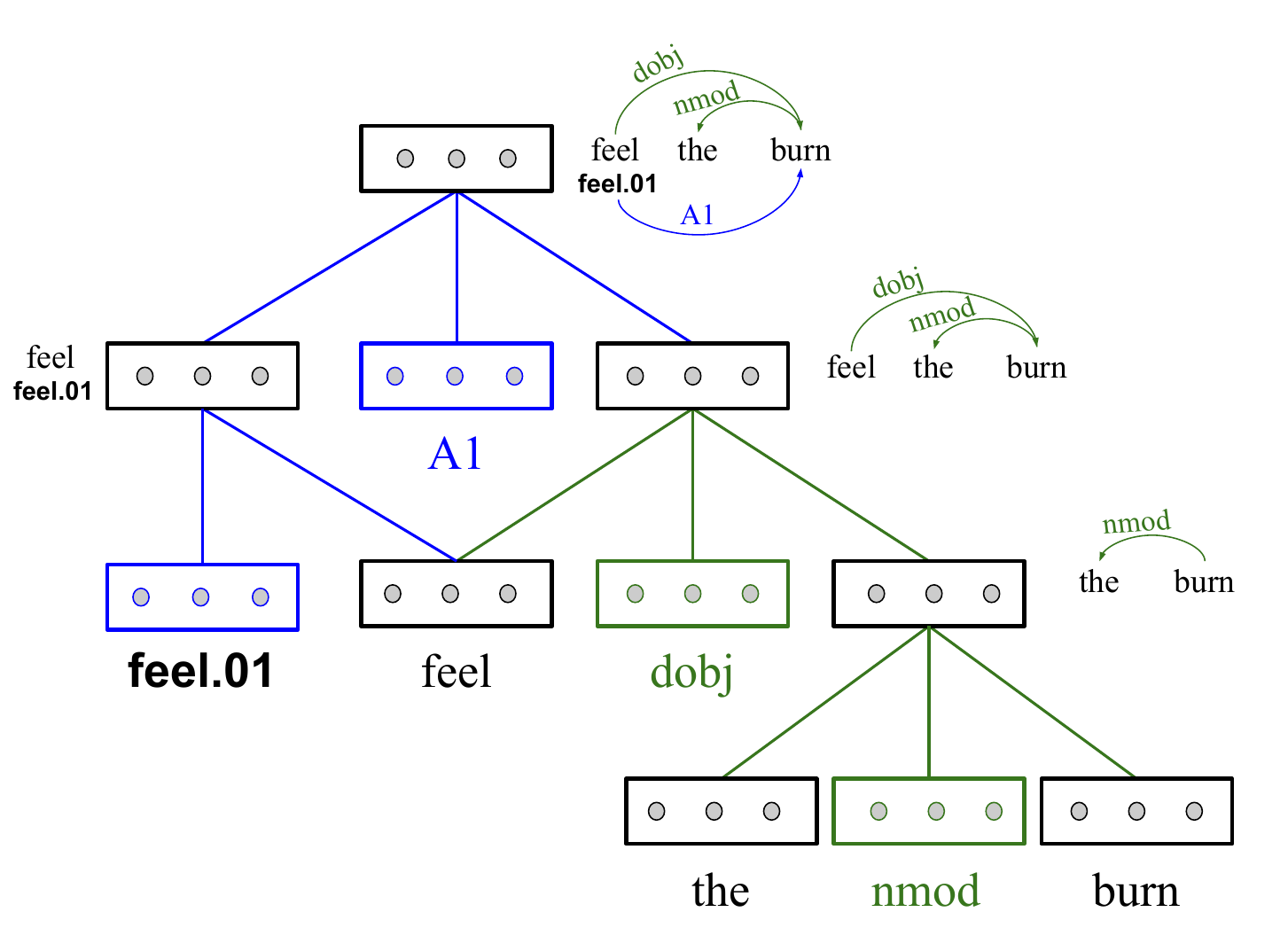}
	\caption{Example of a joint parse tree fragment with vector representations shown at each node. The vectors are obtained by recursive composition of representations of head, dependent, and label vectors. Syntactic dependencies and labels are in green, semantic in blue.\label{fig:comp}}
\end{figure}

Finally, for predicates, we use another recursive function to compose the word representation, $\mathbf{v}$ with a learned representation for the dismabiguated sense of the predicate, $\mathbf{p}$:
\begin{align}
g_{d}(\mathbf{v}, \mathbf{p}) =& \tanh( \mathbf{Z}_d [\mathbf{v}; \mathbf{p}] + \mathbf{e}_d ) 
\end{align}
where $\mathbf{Z}_d$ and $\mathbf{e}_d$ are parameters of the model. Note that, because syntactic and semantic transitions are interleaved, the fragmented structures are a blend of syntactic and semantic compositions. Figure~\ref{fig:comp} shows an example. 

\subsection{Training}
\label{sec:training}

Training the classifier  requires transforming each training instance (a joint parse) into a transition sequence, a deterministic operation under our transition set. Given a collection of algorithm states at time $t$ and correct classification decisions $\tau_t$, we minimize the sum of log-loss terms, given (for one timestep) by:

\begin{align}
- \log \frac{\exp (\mathbf{q}_{\tau_t} + \boldsymbol{\theta}_{\tau_t} \cdot \mathbf{y}_t)}{ \sum_{\tau' \in \mathcal{A}_t} \exp (\mathbf{q}_{\tau'} + \boldsymbol{\theta}_{\tau'} \cdot \mathbf{y}_t)}
\end{align}
with respect to the classifier and LSTM parameters. Note that the loss is \ignore{(sub)}differentiable with respect to the parameters; \ignore{(sub)}gradients are calculated using backpropagation. We apply stochastic \ignore{(sub)}gradient descent with dropout for all neural network parameters.

\subsection{Pretrained Embeddings}
\label{sec:pretrained}

Following~\newcite{Dyer:15}, ``structured skip-gram" embeddings~\cite{Ling:15} were used, trained on the English (AFP section), German, Spanish and Chinese Gigaword corpora, with a window of size 5; training was stopped after 5 epochs. For out-of-vocabulary words, a randomly initialized vector of the same dimension was used.

\subsection{Predicate Sense Disambiguation}
Predicate sense disambiguation is handled within the model (\textsc{M-Pred} transitions), but since senses are lexeme-specific, we need a way to handle unseen predicates at test time. When a predicate is encountered at test time that was not observed in training, our system constructs a predicate from the predicted lemma of the word at that position and defaults to the ``01'' sense, which is correct for 91.22\% of predicates by type in the English CoNLL 2009 training data.

\section{Experimental Setup}
\label{sec:experiments}

Our model is evaluated on the CoNLL shared tasks on joint syntactic and semantic dependency parsing in 2008~\cite{Surdeanu:08} and 2009~\cite{Hajivc:09}. The standard training, development and test splits of all datasets were used. Per the shared task guidelines, automatically predicted POS tags and lemmas provided in the datasets were used for all experiments. As a preprocessing step, pseudo-projectivization of the syntactic trees~\cite{Nivre:07} was used, which allowed an accurate conversion of even the non-projective syntactic trees into syntactic transitions. However, the oracle conversion of semantic parses into transitions is not perfect despite using the \textsc{M-Swap} action, due to the presence of multiple crossing arcs.\footnote{For $1.5\%$ of English sentences in the CoNLL 2009 English dataset, the transition sequence incorrectly encodes the gold-standard joint parse; details in~\newcite{Henderson:13}.}

The standard evaluation metrics include the syntactic labeled attachment score (LAS), the semantic $F_1$ score on both in-domain (WSJ) and out-of-domain (Brown corpus) data, and their macro average (Macro $F_1$) to score joint systems. Because the task was defined somewhat differently in each year, each dataset is considered in turn.

\subsection{CoNLL 2008} 
\label{sec:conll08}

The CoNLL 2008 dataset contains annotations from the Penn Treebank~\cite{Marcus:93}, PropBank~\cite{Palmer:05} and NomBank~\cite{Meyers:04}. The shared task evaluated systems on predicate identification in addition to predicate sense disambiguation and SRL. 

To identify predicates, we trained a zero-Markov order bidirectional LSTM two-class classifier. As input to the classifier, we use learned representations of word lemmas and POS tags. This model achieves an $F_1$ score of 91.43\% on marking words as predicates (or not).

\paragraph{Hyperparameters} The input representation for a word consists of pretrained embeddings (size 100 for English, 80 for Chinese, 64 for German and Spanish), concatenated with additional learned word and POS tag embeddings (size 32 and 12, respectively). Learned embeddings for syntactic and semantic arc labels are of size 20 and predicates 100. Two-layer LSTMs with hidden state dimension 100 were used for each of the four stacks. The parser state $\mathbf{y_t}$ and the composition function $\mathbf{g}$ are of dimension 100. A dropout rate of 0.2~\cite{Zaremba:14} was used on all layers at training time, tuned on the development data from the set of values $\{0.1,0.2,0.3,1.0\}$. The learned representations for actions are of size 100, similarly tuned from $\{10,20,30,40,100\}$. Other hyperparameters have been set intuitively; careful tuning is expected to yield improvements~\cite{Weiss:15}.

An initial learning rate of 0.1 for stochastic gradient descent was used and updated in every training epoch with a decay rate of 0.1~\cite{Dyer:15}. Training is stopped when the development performance does not improve for approximately 6--7 hours of elapsed time. Experiments were run on a single thread on a CPU, with memory requirements of up to 512 MB. 

\subsection{CoNLL 2009} 
\label{sec:CoNLL09}

Relative to the CoNLL 2008 task (above), the main change in 2009 is that predicates are pre-identified, and systems are only evaluated on predicate sense disambiguation (not identification).  Hence, the bidirectional LSTM classifier is not used here. The preprocessing for projectivity, and the hyperparameter selection is the same as in \sect{sec:conll08}.

In addition to the joint approach described in the preceding sections, we experiment here with several variants:

\paragraph{Semantics-only:}  the set of syntactic transitions $\mathcal{S}$, the syntactic stack $S$, and the syntactic composition function $g_s$ are discarded. As a result, the set of constraints on transitions is a subset of the full set of constraints in \sect{sec:constraints}. Effectively, this model does not use any syntactic features, similar to~\newcite{Collobert:11} and~\newcite{Zhou:15}.  It provides a controlled test of the benefit of explicit syntax in a semantic parser.

\paragraph{Syntax-only:} all semantic transitions in $\mathcal{M}$, the semantic stack $M$, and the semantic composition function $g_m$ are discarded. \textsc{S-Shift} and \textsc{S-Right} now \textit{move} the item from the front of the buffer to the syntactic stack, instead of copying. The set of constraints on the transitions is again a subset of the full set of constraints. This model is an arc-eager variant of~\newcite{Dyer:15}, and serves to check whether semantic parsing degrades syntactic performance.

\paragraph{Hybrid:} the semantics parameters are trained using automatically predicted syntax from the syntax-only model. At test time, only semantic parses are predicted. This setup bears similarity to other approaches which pipeline syntax and semantics, extracting features from the syntactic parse to help SRL. However, unlike other approaches, this model does not offer the entire syntactic tree for feature extraction, since only the partial syntactic structures present on the syntactic stack (and potentially the buffer) are visible at a given timestep.  This model helps show the effect of joint prediction.

\section{Results and Discussion}
\label{sec:results}

\paragraph{CoNLL 2008 (\tabr{tab:joint08})} Our joint model significantly outperforms the joint model of~\newcite{Henderson:08}, from which our set of transitions is derived, showing the benefit of learning a representation for the entire algorithmic state. Several other joint learning models have been proposed~\cite{Lluis:08,Johansson:09,Titov:09} for the same task; our joint model surpasses the performance of all these models. The best reported systems on the CoNLL 2008 task are due to~\newcite{Johansson:08},~\newcite{Che:08},~\newcite{Ciaramita:08} and~\newcite{Zhao:08}, all of which pipeline syntax and semantics; our system's semantic and overall performance is comparable to these. We fall behind only~\newcite{Johansson:08}, whose success was attributed to carefully designed global SRL features integrated into a pipeline of classifiers, making them asymptotically slower. 

\begin{table}
	\centering
	\scalebox{0.9}{
		\begin{tabular}{|l|r|r|r|}
			\hline
			\multirow{2}{*}{\bf Model} & \multirow{2}{*}{\bf LAS} & \bf \small Sem. & \bf \small Macro \\
			& & $F_1$ & $F_1$  \\ 
			\hline
			\emph{joint models:} & &  & \\
			\newcite{Lluis:08} & 85.8 & 70.3 & 78.1 \\
			\newcite{Henderson:08} & 87.6 & 73.1 & 80.5  \\ 
			\newcite{Johansson:09} & 86.6 & 77.1 & 81.8  \\ 
			\newcite{Titov:09} & 87.5 & 76.1 & 81.8 \\\hline
			\emph{CoNLL 2008 best:} & & & \\
			\#3: \newcite{Zhao:08} & 87.7 & 76.7 & 82.2 \\
			\#2: \newcite{Che:08} & 86.7 & 78.5 & 82.7  \\ 
			\#2: \newcite{Ciaramita:08} & 87.4 & 78.0 & 82.7  \\ 
			\#1: J\&N (2008) \nocite{Johansson:08} &  89.3 &  81.6 &  85.5  \\ 
			\hline
			Joint (this work) & 89.1 & 80.5 & 84.9  \\ 
			\hline
	\end{tabular}}
	\caption{\label{tab:joint08} Joint parsers: comparison on the CoNLL 2008 test (WSJ+Brown) set.}
\end{table}

\paragraph{CoNLL 2009 English (\tabr{tab:full09})} All of our models (Syntax-only, Semantics-only, Hybrid and Joint) improve over~\newcite{Gesmundo:09} and~\newcite{Henderson:13}, demonstrating the benefit of our entire-parser-state representation learner compared to the more locally scoped model. 

Given that syntax has consistently proven useful in SRL, we expected our Semantics-only model to underperform Hybrid and Joint, and it did.  In the training domain, syntax and semantics benefit each other (Joint outperforms Hybrid).  Out-of-domain (the Brown test set), the Hybrid pulls ahead, a sign that Joint overfits to WSJ. As a syntactic parser, our Syntax-only model performs slightly better than~\newcite{Dyer:15}, who achieve $89.56$ LAS on this task.  Joint parsing is very slightly better still.

The overall performance of Joint is on par with the other winning participants at the CoNLL 2009 shared task~\cite{Zhao:09,Che:09,Gesmundo:09}, falling behind only~\newcite{Zhao:09}, who carefully designed language-specific features  and used a series of pipelines for the joint task, resulting in an accurate but computationally expensive system.

State-of-the-art SRL systems (shown in the last block of \tabr{tab:full09}) which use advances orthogonal to the contributions in this paper, perform better than our models. Many of these systems use expert-crafted features derived from full syntactic parses in a pipeline of classifiers followed by a global reranker~\cite{Bjorkelund:09,Bjorkelund:10,Roth:14}; we have not used these features or reranking.~\newcite{Lei:15} use syntactic parses to obtain interaction features between predicates and their arguments and then compress feature representations using a low-rank tensor. \newcite{Tackstrom:15} present an exact inference algorithm for SRL based on dynamic programming and their local and structured models make use of many syntactic features from a pipeline; our search procedure is greedy. Their algorithm is adopted by~\newcite{Fitzgerald:15} for inference in a model that jointly learns representations from a combination of PropBank and FrameNet annotations; we have not experimented with extra annotations.

Our system achieves an end-to-end runtime of $177.6 \rpm 18$ seconds to parse the CoNLL 2009 English test set on a single core. This is almost 2.5 times faster than the pipeline model of~\newcite{Lei:15} ($439.9 \rpm 42$ seconds) on the same machine.\footnote{See \url{https://github.com/taolei87/SRLParser}; unlike other state-of-the-art systems, this one is publicly available.}

\begin{table}
	\centering
	\scalebox{0.85}{
		\begin{tabular}{|l|r|r|r|r|}
			\cline{1-5}
			\bf Model & \bf LAS & \bf \specialcell{Sem. $F_1$\\(WSJ)} & \bf \specialcell{Sem.~$F_1$\\(Brown)} & \bf \specialcell{Macro \\ $F_1$} \\ \hline
			\emph{CoNLL'09 best:} &&&& \\ 
			\#3 G+ '09 \nocite{Gesmundo:09}  & 88.79 & 83.24 & 70.65 & 86.03 \\
			\#2 C+ '09 \nocite{Che:09} & 88.48 & 85.51 & 73.82 & 87.00  \\ 
			\#1 Z+ '09a \nocite{Zhao:09} & 89.19 & 86.15 & 74.58 & 87.69  \\ \hline
			\emph{this work:} &&&& \\
			Syntax-only & 89.83 &  		&  		&   \\
			Sem.-only  	&  		& 84.39 & 73.87 &   \\ 
			Hybrid 		& 89.83 & 84.58 & 75.64 & 87.20  \\ 
			Joint 		& 89.94 & 84.97 & 74.48 & 87.45  \\\hline
			\emph{pipelines:} &&&& \\
			R\&W '14 \nocite{Roth:14} &  & 86.34 & 75.90 &   \\ 
			L+ '15 \nocite{Lei:15} &  & 86.58 & 75.57 &   \\ 
			T+ '15 \nocite{Tackstrom:15} &  & 87.30 & 75.50 &   \\ 
			F+ '15 \nocite{Fitzgerald:15} &  & 87.80 & 75.50 &   \\ \hline
		\end{tabular}
	}
	\caption{Comparison on the CoNLL 2009 English test set. The first block presents results of other models evaluated for both syntax and semantics on the CoNLL 2009 task. The second block presents our models. The third block presents the best published models, each using its own syntactic preprocessing. 
		\label{tab:full09} }
\end{table}

\paragraph{CoNLL 2009 Multilingual (\tabr{tab:languages})} 
We tested the joint model on the non-English CoNLL 2009 datasets, and the results demonstrate that it adapts easily---it is on par with the top three systems in most cases.  We note that our Chinese parser relies on pretrained word embeddings for its superior performance; without them (not shown), it was on par with the others.  Japanese is a small-data case (4,393 training examples), illustrating our model's dependence on reasonably large training datasets. 

We have not extended our model to incorporate morphological features, which are used by the systems to which we compare.  Future work might incorporate morphological features where available; this could potentially improve performance, especially in highly inflective languages like Czech.  An alternative might be to infer word-internal representations using character-based word embeddings, which was found beneficial for syntactic parsing \cite{Ballesteros:15}.

\begin{table}[ht]
	\centering
	\scalebox{0.83}{
		\begin{tabular}{|l|r|r|r|r|}
			\hline
			Language 	&  \#1 {C+'09}	& \#2 {Z+ '09a} & \#3 {G+ '09} 	& Joint \\
			\hline
			Catalan		& 81.84			& \bf 83.01 	& 82.66			& 82.40 \\
			Chinese		& 76.38			& 76.23 		& 76.15			& \bf 79.27\\
			Czech		& \bf 83.27		& 80.87			& 83.21			& 79.53 \\
			English		& 87.00			& \bf 87.69		& 86.03			& 87.45 \\
			German		& \bf 82.44		& 81.22 		& 79.59			& 81.05 \\
			Japanese	& \bf 85.65		& 85.28			& 84.91			& 80.91 \\
			Spanish		& 81.90			& \bf 83.31		& 82.43 		& 83.11 \\
			\hline
			Average		& 82.64			& 82.52 		& 82.14			& 81.96 \\
			\hline
	\end{tabular}}
	\caption{\label{tab:languages} Comparison of macro $F_1$ scores on the multilingual CoNLL 2009 test set.}
\end{table}

\section{Related Work}
\label{sec:related} 

Other approaches to joint modeling, not considered in our experiments, are notable.~\newcite{Lluis:13} propose a graph-based joint model using dual decomposition for agreement between syntax and semantics, but do not achieve competitive performance on the CoNLL 2009 task.~\newcite{Lewis:15} proposed an efficient joint model for CCG syntax and SRL, which performs better than a pipelined model. However, their training necessitates CCG annotation, ours does not. Moreover, their evaluation metric rewards semantic dependencies regardless of where they attach within the argument span given by a PropBank constituent, making direct comparison to our evaluation  infeasible. \newcite{Krishnamurthy:14} propose a joint CCG parsing and relation extraction model which improves over pipelines, but their task is different from ours.~\newcite{Li:10} also perform joint syntactic and semantic dependency parsing for Chinese, but do not report results on the CoNLL 2009 dataset.

There has also been an increased interest in models which use neural networks for SRL.~\newcite{Collobert:11} proposed models which perform many NLP tasks without hand-crafted features. Though they did not achieve the best results on the constituent-based SRL task~\cite{Carreras:05}, their approach inspired~\newcite{Zhou:15}, who achieved state-of-the-art results using deep bidirectional LSTMs. 
Our approach for dependency-based SRL is not directly comparable.

\section{Conclusion}
\label{sec:conclusion}

We presented an incremental, greedy parser for joint syntactic and semantic dependency parsing. Our model surpasses the performance of previous joint models on the CoNLL 2008 and 2009 English tasks, without using expert-crafted, expensive features of the full syntactic parse.

\section*{Acknowledgments}
The authors thank Sam Thomson, Lingpeng Kong, Mark Yatskar, Eunsol Choi, George Mulcaire, and Luheng He, as well as the anonymous reviewers, for many useful comments. This research was supported in part by DARPA grant FA8750-12-2-0342 funded under the DEFT program and by the U.S.~Army Research Office under grant number W911NF-10-1-0533.  Any opinion, findings, and conclusions or recommendations expressed in this material are those of the author(s) and do not necessarily reflect the view of the U.S.~Army Research Office or the U.S.~Government. Miguel Ballesteros was supported by the European Commission under the contract numbers FP7-ICT-610411 (project MULTISENSOR) and H2020-RIA-645012 (project KRISTINA).

\bibliographystyle{acl2016}
\bibliography{acl2016}

\begin{thebibliography}{}

\bibitem[\protect\citename{Ballesteros and Nivre}2013]{Ballesteros:13}
Miguel Ballesteros and Joakim Nivre.
\newblock 2013.
\newblock Going to the roots of dependency parsing.
\newblock {\em Computational Linguistics}, 39(1):5--13.

\bibitem[\protect\citename{Ballesteros \bgroup et al.\egroup
  }2015]{Ballesteros:15}
Miguel Ballesteros, Chris Dyer, and Noah~A. Smith.
\newblock 2015.
\newblock Improved transition-based parsing by modeling characters instead of
  words with {LSTM}s.
\newblock In {\em Proc. of EMNLP}.

\bibitem[\protect\citename{Bj\"{o}rkelund \bgroup et al.\egroup
  }2009]{Bjorkelund:09}
Anders Bj\"{o}rkelund, Love Hafdell, and Pierre Nugues.
\newblock 2009.
\newblock Multilingual semantic role labeling.
\newblock In {\em Proc. of CoNLL}.

\bibitem[\protect\citename{Bj{\"o}rkelund \bgroup et al.\egroup
  }2010]{Bjorkelund:10}
Anders Bj{\"o}rkelund, Bernd Bohnet, Love Hafdell, and Pierre Nugues.
\newblock 2010.
\newblock A high-performance syntactic and semantic dependency parser.
\newblock In {\em Proc. of COLING}.

\bibitem[\protect\citename{Carreras and M{\`a}rquez}2005]{Carreras:05}
Xavier Carreras and Llu{\'\i}s M{\`a}rquez.
\newblock 2005.
\newblock Introduction to the {CoNLL}-2005 shared task: Semantic role labeling.
\newblock In {\em Proc. of CoNLL}.

\bibitem[\protect\citename{Che \bgroup et al.\egroup }2008]{Che:08}
Wanxiang Che, Zhenghua Li, Yuxuan Hu, Yongqiang Li, Bing Qin, Ting Liu, and
  Sheng Li.
\newblock 2008.
\newblock A cascaded syntactic and semantic dependency parsing system.
\newblock In {\em Proc. of CoNLL}.

\bibitem[\protect\citename{Che \bgroup et al.\egroup }2009]{Che:09}
Wanxiang Che, Zhenghua Li, Yongqiang Li, Yuhang Guo, Bing Qin, and Ting Liu.
\newblock 2009.
\newblock Multilingual dependency-based syntactic and semantic parsing.
\newblock In {\em Proc. of CoNLL}.

\bibitem[\protect\citename{Ciaramita \bgroup et al.\egroup }2008]{Ciaramita:08}
Massimiliano Ciaramita, Giuseppe Attardi, Felice Dell'Orletta, and Mihai
  Surdeanu.
\newblock 2008.
\newblock {DeSRL}: A linear-time semantic role labeling system.
\newblock In {\em Proc. of CoNLL}.

\bibitem[\protect\citename{Collobert \bgroup et al.\egroup }2011]{Collobert:11}
Ronan Collobert, Jason Weston, L{\'e}on Bottou, Michael Karlen, Koray
  Kavukcuoglu, and Pavel Kuksa.
\newblock 2011.
\newblock Natural language processing (almost) from scratch.
\newblock {\em Journal of Machine Learning Research}, 12:2493--2537.

\bibitem[\protect\citename{Dyer \bgroup et al.\egroup }2015]{Dyer:15}
Chris Dyer, Miguel Ballesteros, Wang Ling, Austin Matthews, and Noah~A. Smith.
\newblock 2015.
\newblock Transition-based dependency parsing with stack long short-term
  memory.
\newblock In {\em Proc. of ACL}.

\bibitem[\protect\citename{FitzGerald \bgroup et al.\egroup
  }2015]{Fitzgerald:15}
Nicholas FitzGerald, Oscar T{\"{a}}ckstr{\"{o}}m, Kuzman Ganchev, and Dipanjan
  Das.
\newblock 2015.
\newblock Semantic role labelling with neural network factors.
\newblock In {\em Proc. of EMNLP}.

\bibitem[\protect\citename{Foland and Martin}2015]{Foland:15}
William~R. Foland and James Martin.
\newblock 2015.
\newblock Dependencybased semantic role labeling using convolutional neural
  networks.
\newblock In {\em Proc. of *SEM}.

\bibitem[\protect\citename{Gesmundo \bgroup et al.\egroup }2009]{Gesmundo:09}
Andrea Gesmundo, James Henderson, Paola Merlo, and Ivan Titov.
\newblock 2009.
\newblock A latent variable model of synchronous syntactic-semantic parsing for
  multiple languages.
\newblock In {\em Proc. of CoNLL}.

\bibitem[\protect\citename{Gildea and Jurafsky}2002]{Gildea:02a}
Daniel Gildea and Daniel Jurafsky.
\newblock 2002.
\newblock Automatic labeling of semantic roles.
\newblock {\em Computational Linguistics}, 28(3):245--288.

\bibitem[\protect\citename{Goldberg}2015]{Goldberg:15}
Yoav Goldberg.
\newblock 2015.
\newblock A primer on neural network models for natural language processing.
\newblock arXiv:1510.00726.

\bibitem[\protect\citename{Graves}2013]{Graves:13}
Alex Graves.
\newblock 2013.
\newblock Generating sequences with recurrent neural networks.
\newblock arXiv:1308.0850.

\bibitem[\protect\citename{Haji\v{c} \bgroup et al.\egroup }2009]{Hajivc:09}
Jan Haji\v{c}, Massimiliano Ciaramita, Richard Johansson, Daisuke Kawahara,
  Maria~Ant\`{o}nia Mart\'{\i}, Llu\'{i}s M\`{a}rquez, Adam Meyers, Joakim
  Nivre, Sebastian Pad\'{o}, Jan \v{S}t\v{e}p\'{a}nek, Pavel Stra\v{n}\'{a}k,
  Mihai Surdeanu, Nianwen Xue, and Yi~Zhang.
\newblock 2009.
\newblock The {CoNLL}-2009 shared task: Syntactic and semantic dependencies in
  multiple languages.
\newblock In {\em Proc. of CoNLL}.

\bibitem[\protect\citename{He \bgroup et al.\egroup }2013]{He:13}
He~He, Hal Daum{\'e}~III, and Jason Eisner.
\newblock 2013.
\newblock Dynamic feature selection for dependency parsing.
\newblock In {\em Proc. of EMNLP}.

\bibitem[\protect\citename{Henderson \bgroup et al.\egroup }2008]{Henderson:08}
James Henderson, Paola Merlo, Gabriele Musillo, and Ivan Titov.
\newblock 2008.
\newblock A latent variable model of synchronous parsing for syntactic and
  semantic dependencies.
\newblock In {\em Proc. of CoNLL}.

\bibitem[\protect\citename{Henderson \bgroup et al.\egroup }2013]{Henderson:13}
James Henderson, Paola Merlo, Ivan Titov, and Gabriele Musillo.
\newblock 2013.
\newblock Multi-lingual joint parsing of syntactic and semantic dependencies
  with a latent variable model.
\newblock {\em Computational Linguistics}, 39(4):949--998.

\bibitem[\protect\citename{Hochreiter and Schmidhuber}1997]{Hochreiter:97}
Sepp Hochreiter and J{\"{u}}rgen Schmidhuber.
\newblock 1997.
\newblock Long short-term memory.
\newblock {\em Neural Computation}, 9(8):1735--1780.

\bibitem[\protect\citename{Johansson and Nugues}2008]{Johansson:08}
Richard Johansson and Pierre Nugues.
\newblock 2008.
\newblock Dependency-based syntactic-semantic analysis with {PropBank} and
  {NomBank}.
\newblock In {\em Proc. of CoNLL}.

\bibitem[\protect\citename{Johansson}2009]{Johansson:09}
Richard Johansson.
\newblock 2009.
\newblock Statistical bistratal dependency parsing.
\newblock In {\em Proc. of EMNLP}.

\bibitem[\protect\citename{Krishnamurthy and Mitchell}2014]{Krishnamurthy:14}
Jayant Krishnamurthy and Tom~M. Mitchell.
\newblock 2014.
\newblock Joint syntactic and semantic parsing with combinatory categorial
  grammar.
\newblock In {\em Proc. of ACL}.

\bibitem[\protect\citename{Lei \bgroup et al.\egroup }2015]{Lei:15}
Tao Lei, Yuan Zhang, Llu{\'{\i}}s~M{\`{a}}rquez i~Villodre, Alessandro
  Moschitti, and Regina Barzilay.
\newblock 2015.
\newblock High-order low-rank tensors for semantic role labeling.
\newblock In {\em Proc. of NAACL}.

\bibitem[\protect\citename{Lewis \bgroup et al.\egroup }2015]{Lewis:15}
Mike Lewis, Luheng He, and Luke Zettlemoyer.
\newblock 2015.
\newblock Joint {A* CCG} parsing and semantic role labelling.
\newblock In {\em Proc. of EMNLP}.

\bibitem[\protect\citename{Li \bgroup et al.\egroup }2010]{Li:10}
Junhui Li, Guodong Zhou, and Hwee~Tou Ng.
\newblock 2010.
\newblock Joint syntactic and semantic parsing of {Chinese}.
\newblock In {\em Proc. of ACL}.

\bibitem[\protect\citename{Ling \bgroup et al.\egroup }2015]{Ling:15}
Wang Ling, Chris Dyer, Alan Black, and Isabel Trancoso.
\newblock 2015.
\newblock Two/too simple adaptations of word2vec for syntax problems.
\newblock In {\em Proc. of NAACL}.

\bibitem[\protect\citename{Llu\'{\i}s and M\`{a}rquez}2008]{Lluis:08}
Xavier Llu\'{\i}s and Llu\'{\i}s M\`{a}rquez.
\newblock 2008.
\newblock A joint model for parsing syntactic and semantic dependencies.
\newblock In {\em Proc. of CoNLL}.

\bibitem[\protect\citename{Llu{\'\i}s \bgroup et al.\egroup }2013]{Lluis:13}
Xavier Llu{\'\i}s, Xavier Carreras, and Llu{\'\i}s M{\`a}rquez.
\newblock 2013.
\newblock Joint arc-factored parsing of syntactic and semantic dependencies.
\newblock {\em Transactions of the ACL}, 1:219--230.

\bibitem[\protect\citename{Marcus \bgroup et al.\egroup }1993]{Marcus:93}
Mitchell~P. Marcus, Mary~Ann Marcinkiewicz, and Beatrice Santorini.
\newblock 1993.
\newblock Building a large annotated corpus of {English}: The {Penn} treebank.
\newblock {\em Computational Linguistics}, 19(2):313--330.

\bibitem[\protect\citename{Meyers \bgroup et al.\egroup }2004]{Meyers:04}
Adam Meyers, Ruth Reeves, Catherine Macleod, Rachel Szekely, Veronika
  Zielinska, Brian Young, and Ralph Grishman.
\newblock 2004.
\newblock The {NomBank} project: An interim report.
\newblock In {\em Proc. of NAACL}.

\bibitem[\protect\citename{Nair and Hinton}2010]{Nair:10}
Vinod Nair and Geoffrey~E. Hinton.
\newblock 2010.
\newblock Rectified linear units improve restricted {Boltzmann} machines.
\newblock In {\em Proc. of ICML}.

\bibitem[\protect\citename{Nivre \bgroup et al.\egroup }2007]{Nivre:07}
Joakim Nivre, Johan Hall, Jens Nilsson, Atanas Chanev, G{\"{u}}lsen Eryigit,
  Sandra K{\"{u}}bler, Svetoslav Marinov, and Erwin Marsi.
\newblock 2007.
\newblock {MaltParser}: A language-independent system for data-driven
  dependency parsing.
\newblock {\em Natural Language Engineering}, 13:95--135.

\bibitem[\protect\citename{Nivre}2008]{Nivre:08}
Joakim Nivre.
\newblock 2008.
\newblock Algorithms for deterministic incremental dependency parsing.
\newblock {\em Computational Linguistics}, 34(4):513--553.

\bibitem[\protect\citename{Nivre}2009]{Nivre:09}
Joakim Nivre.
\newblock 2009.
\newblock Non-projective dependency parsing in expected linear time.
\newblock In {\em Proc. of ACL}.

\bibitem[\protect\citename{Palmer \bgroup et al.\egroup }2005]{Palmer:05}
Martha Palmer, Daniel Gildea, and Paul Kingsbury.
\newblock 2005.
\newblock The {Proposition Bank}: An annotated corpus of semantic roles.
\newblock {\em Computational Linguistics}, 31(1):71--106.

\bibitem[\protect\citename{Roth and Woodsend}2014]{Roth:14}
Michael Roth and Kristian Woodsend.
\newblock 2014.
\newblock Composition of word representations improves semantic role labelling.
\newblock In {\em Proc. of EMNLP}.

\bibitem[\protect\citename{Surdeanu \bgroup et al.\egroup }2008]{Surdeanu:08}
Mihai Surdeanu, Richard Johansson, Adam Meyers, Llu\'{\i}s M\`{a}rquez, and
  Joakim Nivre.
\newblock 2008.
\newblock The {CoNLL}-2008 shared task on joint parsing of syntactic and
  semantic dependencies.
\newblock In {\em Proc. of CoNLL}.

\bibitem[\protect\citename{Sutton and McCallum}2005]{Sutton:05}
Charles Sutton and Andrew McCallum.
\newblock 2005.
\newblock Joint parsing and semantic role labeling.
\newblock In {\em Proc. of CoNLL}.

\bibitem[\protect\citename{T{\"{a}}ckstr{\"{o}}m \bgroup et al.\egroup
  }2015]{Tackstrom:15}
Oscar T{\"{a}}ckstr{\"{o}}m, Kuzman Ganchev, and Dipanjan Das.
\newblock 2015.
\newblock Efficient inference and structured learning for semantic role
  labeling.
\newblock {\em Transactions of the ACL}, 3:29--41.

\bibitem[\protect\citename{Titov \bgroup et al.\egroup }2009]{Titov:09}
Ivan Titov, James Henderson, Paola Merlo, and Gabriele Musillo.
\newblock 2009.
\newblock Online graph planarisation for synchronous parsing of semantic and
  syntactic dependencies.
\newblock In {\em Proc. of IJCAI}.

\bibitem[\protect\citename{Toutanova \bgroup et al.\egroup }2008]{Toutanova:08}
Kristina Toutanova, Aria Haghighi, and Christopher~D. Manning.
\newblock 2008.
\newblock A global joint model for semantic role labeling.
\newblock {\em Computational Linguistics}, 34(2):161--191.

\bibitem[\protect\citename{Weiss \bgroup et al.\egroup }2015]{Weiss:15}
David Weiss, Chris Alberti, Michael Collins, and Slav Petrov.
\newblock 2015.
\newblock Structured training for neural network transition-based parsing.
\newblock In {\em Proc. of ACL}.

\bibitem[\protect\citename{Zaremba \bgroup et al.\egroup }2014]{Zaremba:14}
Wojciech Zaremba, Ilya Sutskever, and Oriol Vinyals.
\newblock 2014.
\newblock Recurrent neural network regularization.
\newblock arXiv:1409.2329.

\bibitem[\protect\citename{Zhao and Kit}2008]{Zhao:08}
Hai Zhao and Chunyu Kit.
\newblock 2008.
\newblock Parsing syntactic and semantic dependencies with two single-stage
  maximum entropy models.
\newblock In {\em Proc. of CoNLL}.

\bibitem[\protect\citename{Zhao \bgroup et al.\egroup }2009]{Zhao:09}
Hai Zhao, Wenliang Chen, Jun'ichi Kazama, Kiyotaka Uchimoto, and Kentaro
  Torisawa.
\newblock 2009.
\newblock Multilingual dependency learning: Exploiting rich features for
  tagging syntactic and semantic dependencies.
\newblock In {\em Proc. of CoNLL}.

\bibitem[\protect\citename{Zhou and Xu}2015]{Zhou:15}
Jie Zhou and Wei Xu.
\newblock 2015.
\newblock End-to-end learning of semantic role labeling using recurrent neural
  networks.
\newblock In {\em Proc. of ACL}.

\end{thebibliography}
\end{document}